\pdfoutput=1
\documentclass[11pt]{article}

\usepackage[review]{acl}

\usepackage{times}
\usepackage{latexsym}
\usepackage{amsmath}
\usepackage[T1]{fontenc}
\usepackage[utf8]{inputenc}

\usepackage{microtype}

\usepackage{inconsolata}

\usepackage{graphicx}

\title{Catastrophic Forgetting in LLMs: A Comparative Analysis Across Language Tasks}

\author{Naimul Haque \\
  Alfred University \\
  New York, USA \\
  \texttt{naimul011@gmail.com} }

\begin{document}
\maketitle
\section*{Abstract}

Large Language Models (LLMs) have significantly advanced Natural Language Processing (NLP), particularly in Natural Language Understanding (NLU) tasks. As we progress toward an agentic world where LLM-based agents autonomously handle specialized tasks, it becomes crucial for these models to adapt to new tasks without forgetting previously learned information—a challenge known as catastrophic forgetting. This study evaluates the continual fine-tuning of various open-source LLMs with different parameter sizes (specifically models under 10 billion parameters) on key NLU tasks from the GLUE benchmark, including SST-2, MRPC, CoLA, and MNLI. By employing prompt engineering and task-specific adjustments, we assess and compare the models' abilities to retain prior knowledge while learning new tasks. Our results indicate that models such as \textbf{Phi-3.5-mini} exhibit minimal forgetting while maintaining strong learning capabilities, making them well-suited for continual learning environments. Additionally, models like \textbf{Orca-2-7b} and \textbf{Qwen2.5-7B} demonstrate impressive learning abilities and overall performance after fine-tuning. This work contributes to understanding catastrophic forgetting in LLMs and highlights prompting engineering to optimize model performance for continual learning scenarios.

\section{Introduction}

Large Language Models (LLMs) \cite{attention} have transformed Natural Language Processing (NLP), delivering state-of-the-art performance on various tasks like sentiment analysis, paraphrase detection, and natural language inference \cite{bubeck2023sparks}. Open-source models such as \textbf{Llama 3} \cite{llama3.2} have become essential in advancing Natural Language Understanding (NLU) \cite{glue} tasks, which are critical for interpreting and generating human language.

As we move towards an \textit{agentic world} \cite{agents} where LLM-based agents autonomously handle specialized tasks, the ability to fine-tune models on multiple tasks without losing accuracy or forgetting previously learned information is crucial. Continual fine-tuning (CF) \cite{continual} enables models to adapt and excel in varied environments while maintaining performance across tasks.

While prior research has focused on text generation capabilities \cite{empiricalstudycatastrophicforgetting}, less attention has been given to addressing CF in NLU tasks \cite{glue}, which are vital for real-world applications. Furthermore, there has been limited exploration of how models with different parameter sizes handle CF during continual fine-tuning, especially models under 10 billion parameters.

This study aims to fill these gaps by evaluating forgetting across various models with different parameter sizes on key NLU tasks from the GLUE benchmark \cite{glue}, including SST-2, MRPC, CoLA, and MNLI. By conducting extensive experimentation with prompt engineering and task-specific adjustments, we provide comparative insights into how models like \textbf{Orca-2-7b} \cite{orca}, \textbf{Llama-3.1-8B} \cite{llama3.2}, and \textbf{Phi-3.5-mini} \cite{phi} perform under sequential fine-tuning.

Our results show that models such as \textbf{Phi-3.5-mini} \cite{phi} continue to exhibit minimal forgetting while maintaining strong learning capabilities, making them ideal for continual learning environments. Similarly, \textbf{Orca-2-7b} \cite{orca} and \textbf{Qwen2.5-7B} \cite{qwen25} demonstrate impressive learning abilities and overall performance after fine-tuning. 

In summary, this work contributes by:
\begin{itemize}
    \item Evaluating catastrophic forgetting and task learning across diverse LLMs using sequential fine-tuning on specific NLU tasks.
    \item Highlighting the importance of prompt engineering and fine-tuning strategies for optimizing model performance.
    \item Providing insights into the performance of models with different parameter sizes, identifying those best suited for continual learning.
    \item Proposing continual fine-tuning as a key strategy for future LLM agents to handle multiple tasks without sacrificing accuracy.
\end{itemize}

%The templates include the \LaTeX{} source of this document (\texttt{acl\_latex.tex}),
%the \LaTeX{} style file used to format it (\texttt{acl.sty}),
%an ACL bibliography style (\texttt{acl\_natbib.bst}),
%an example bibliography (\texttt{custom.bib}),
%and the bibliography for the ACL Anthology (\texttt{anthology.bib}).

\section{Related Works}
Catastrophic forgetting, a phenomenon first identified by \cite{mccloskey1989catastrophic}, remains a fundamental challenge in sequential learning tasks for neural networks. When models are trained on multiple tasks in sequence, they tend to overwrite previously acquired knowledge, leading to significant performance degradation on earlier tasks. Various methods have been proposed to mitigate this issue, such as Elastic Weight Consolidation (EWC) \cite{kirkpatrick2017overcoming}, which regularizes weight updates to protect crucial parameters learned from previous tasks. Similarly, memory-based methods like Gradient Episodic Memory (GEM) \cite{lopez2017gradient} address forgetting by storing and replaying examples from past tasks during training, thereby reducing interference.

While recent LLM models show strong zero-shot performance, they often struggle with tasks outside their training and evaluation sets. To address this, \cite{tuhin} propose Continual-T0 (CT0), a fine-tuned LLM capable of learning new tasks while retaining prior knowledge, largely due to the self-supervision pre-training process. \cite{kemker2017measuringcatastrophicforgettingneural} showed that Deep neural networks struggle to learn new tasks without forgetting old ones, and various methods have been proposed to mitigate this, but their effectiveness varies depending on the training paradigm and data type. \cite{catastrophicforgettinglarge} conducted an empirical study on catastrophic forgetting in LLMs, finding that forgetting becomes more severe as model size increases, especially in models ranging from 1B to 7B parameters, during continual fine-tuning across domains like reasoning and reading comprehension.

\section{Methodology}
\begin{figure}[t]
 \includegraphics[width=\columnwidth]{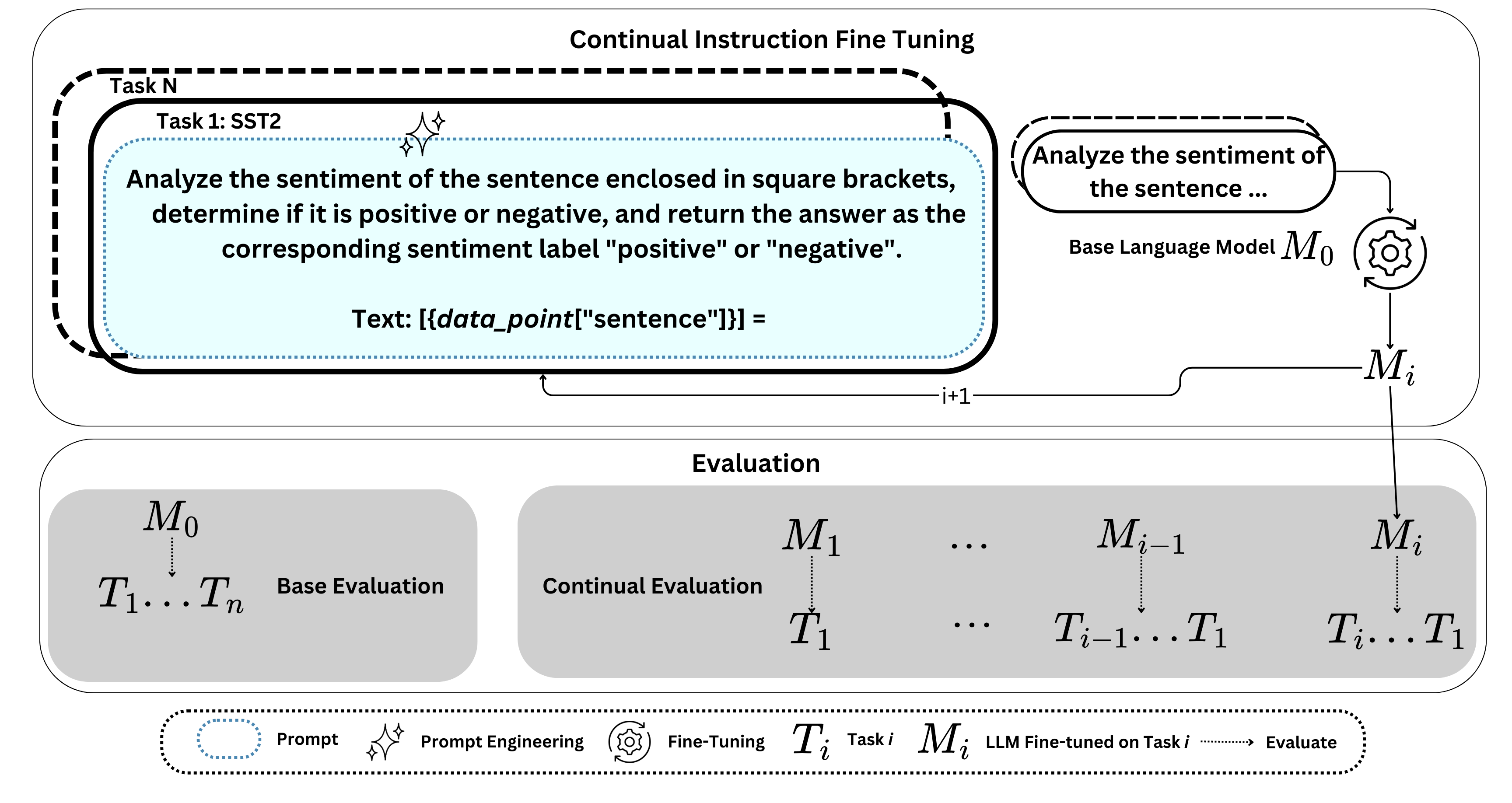}
  \caption{The figure illustrates the Continual Finetuning workflow. $ M_0$ represents the base language model, and subsequent models $ M_1, M_2, \dots, M_n$ denote the fine-tuned versions after training on tasks $ T_1, T_2, \dots, T_n$. The figure also highlights the process of generating task-specific prompts and the continual evaluation to assess the model's retention.}
  \label{fig:method}
\end{figure}

We employed a \textit{Continual Instruction Fine-tuning} \cite{catastrophicforgettinglarge} approach, shown in the Figure~\ref{fig:method}, sequentially adapting the base model $M_0$ on tasks $\{ T_1, T_2, \dots, T_n \}$ from the GLUE benchmark \cite{glue}. The goal was to evaluate how well the model $M_i$, fine-tuned on task $T_i$, retained knowledge from previous tasks $\{ T_1, \dots, T_{i-1} \}$.

The methodology comprised two main stages. First, we prepared the dataset for each task $T_i$ using \textit{prompt engineering} (PE) \cite{promptengineering}. Let $X$ be the original dataset for $T_i$, transformed into a structured prompt dataset $X'$ as follows:

\[
X' = \text{PE}(X)
\]

where $\text{PE}(\cdot)$ represents the prompt engineering function. Prompts were designed to guide the model to perform task-specific instructions and the exact prompts and their expected outputs are shown in Table~\ref{tab:instruction_finetuning}

\begin{table*}[ht]
  \centering
  \resizebox{\textwidth}{!}{
\begin{tabular}{|c|p{15cm}|p{5cm}|}
\hline
\textbf{Task} & \textbf{Instruction} & \textbf{Output Format} \\ 
\hline
SST-2 & Analyze the sentiment of the sentence enclosed in square brackets, determine if it is positive or negative, and return the answer as the corresponding sentiment label "positive" or "negative". & "positive" or "negative" \\ 
\hline
MRPC & Analyze the two sentences enclosed in square brackets and determine if they are paraphrases of each other. & "paraphrase" or "not paraphrase" \\
\hline
CoLA & Determine if the sentence enclosed in square brackets is grammatically correct or incorrect. & "correct" or "incorrect" \\
\hline
MNLI & Analyze the relationship between the premise and hypothesis enclosed in square brackets. Determine if the hypothesis is entailed by, neutral to, or contradicts the premise. & "entailment", "neutral", or "contradiction" \\
\hline
\end{tabular}
}
\caption{
This table presents the instructions used for fine-tuning models on specific tasks such as sentiment analysis (SST-2), paraphrase detection (MRPC), grammatical acceptability (CoLA), and natural language inference (MNLI). }
\label{tab:instruction_finetuning}
\end{table*}

\section{Evaluation and Measurement of Catastrophic Forgetting and Learning}

After each fine-tuning episode, model $M_i$ was evaluated on all previous tasks $\{T_1, T_2, \dots, T_i\}$ to assess catastrophic forgetting and learning. Accuracy was used as the evaluation metric, comparing post-fine-tuning performance to the base performance to detect any degradation or improvement. Forgetting was quantified as the difference between the maximum accuracy on each task during fine-tuning and the final accuracy:

\[
\text{Forgetting for Task } t = \max_{0 \leq k \leq T} (a_{k,t}) - a_{T,t}
\]

where $a_{k,t}$ is the accuracy after fine-tuning on task $k$ and $a_{T,t}$ is the final accuracy. 

Learning was calculated as the improvement over the base performance:

\[
\text{Learning for Task } t = \max_{k \leq T} (a_{k,t}) - a_{0,t}
\]

where $a_{0,t}$ is the base accuracy and $\max_{k \leq T} (a_{k,t})$ is the highest accuracy during fine-tuning. Both metrics provide insight into the model’s behavior across tasks.

\subsection{Selected Tasks}

The selected tasks include:
\begin{enumerate}
    \item \textbf{SST-2} (Stanford Sentiment Treebank): A binary classification task for determining whether a sentence has a positive or negative sentiment.
    \item \textbf{MRPC} (Microsoft Research Paraphrase Corpus): A binary classification task to predict whether two sentences are paraphrases of each other.
    \item \textbf{CoLA} (Corpus of Linguistic Acceptability): A binary classification task for determining whether a sentence is grammatically correct.
    \item \textbf{MNLI} (Multi-Genre Natural Language Inference): A three-class classification task where the goal is to determine whether a premise sentence entails, contradicts, or is neutral with respect to a hypothesis sentence across multiple genres of text.
\end{enumerate}

\section{Experimental Results}

Table~\ref{tab:model_performance} presents a comparison of various models' performance across four key metrics: Pretrained Performance, Forgetting, Learning, and overall Training Performance after sequential task fine-tuning.
\begin{table*}[ht]
  \centering
  \resizebox{\textwidth}{!}{
    \begin{tabular}{lccccc}
      \hline
      \textbf{Model} & \textbf{Parameter Size (B)} & \textbf{Pretrained Performance} & \textbf{Forgetting} & \textbf{Learning} & \textbf{Training Performance} \\
      \hline
      Llama-3.2-1B  \cite{llama3.2}       & 1     & 0.50          & 0.24           & 0.33             & 0.54                  \\
      Llama-3.2-3B   \cite{llama3.2}      & 3     & 0.56          & 0.225          & 0.36             & 0.61                  \\
      Llama-3.1-8B  \cite{llama3.2}       & 8     & 0.56          & 0.59           & 0.84             & 0.67                  \\
      Llama-3-8B   \cite{llama3.2}   & 8     & 0.53          & 0.39           & 0.98             & 0.70                  \\
      Llama-2-7B   \cite{llama3.2}     & 7     & 0.67          & 0.23           & 0.12             & 0.63                  \\
      GPT-J-6B      \cite{mesh-transformer-jax}       & 6     & 0.50          & 0.39           & 0.45             & 0.54                  \\
      Phi-2        \cite{phi}        & 2.7   & 0.59          & 0.1            & 0.15             & 0.61                  \\
      Phi-3.5-mini   \cite{phi}      & 3.82  & 0.69          & \textbf{0.02}  & 0.30             & 0.76                  \\
      Orca-2-7b   \cite{orca}         & 7     & \textbf{0.76} & 0.185          & 0.33             & \textbf{0.81}         \\
      Qwen2.5-0.5B    \cite{qwen25}     & 0.5   & 0.52          & 0.23           & 0.56             & 0.61                  \\
      Qwen2.5-7B     \cite{qwen25}      & 7     & 0.56          & 0.51           & \textbf{1.12}    & 0.77                  \\
      Qwen2.5-14B    \cite{qwen25}      & 14    & 0.71          & 0.935          & 0.66             & 0.80                  \\
      \hline
    \end{tabular}
  }
  \caption{Comparison of the pre-trained performance, forgetting, learning capabilities, and training performance of various language models. \textbf{Parameter Size (B)} indicates the number of parameters (in billions) of each model. The \textbf{Pretrained Performance} refers to the initial performance of each model before any task-specific training. }
  %\textbf{Forgetting} quantifies how much previously learned information is lost after sequential fine-tuning (lower is better). \textbf{Learning} captures the model's ability to improve its performance after additional training (higher is better). The final column, \textbf{Training Performance}, indicates the overall result after the models have been trained on multiple tasks. The best values in each category are highlighted in bold to emphasize superior performance across different metrics.
  \label{tab:model_performance}
\end{table*}

\textbf{Orca-2-7b} \cite{orca} achieved the highest pretraining performance at \textbf{0.76}, demonstrating strong initial capabilities, closely followed by \textbf{Qwen2.5-14B} \cite{qwen25} at \textbf{0.71}, indicating a solid foundation before fine-tuning. In terms of catastrophic forgetting, \textbf{Phi-3.5-mini} \cite{phi} and \textbf{Phi-2} showed minimal forgetting with values of \textbf{0.02} and \textbf{0.1}, respectively. On the other hand, \textbf{Qwen2.5-14B} and \textbf{Llama-3.1-8B} \cite{llama3.2} exhibited high forgetting rates of \textbf{0.935} and \textbf{0.59}. After fine-tuning, \textbf{Orca-2-7b} stood out with the best average performance of \textbf{0.81}, followed by \textbf{Qwen2.5-14B} at \textbf{0.80}. Models like \textbf{Phi-3.5-mini} and \textbf{Orca-2-7b} performed better overall, balancing low forgetting and high learning rates, offering valuable insights into mitigating catastrophic forgetting.

\begin{figure}[t]
  \includegraphics[width=\columnwidth]{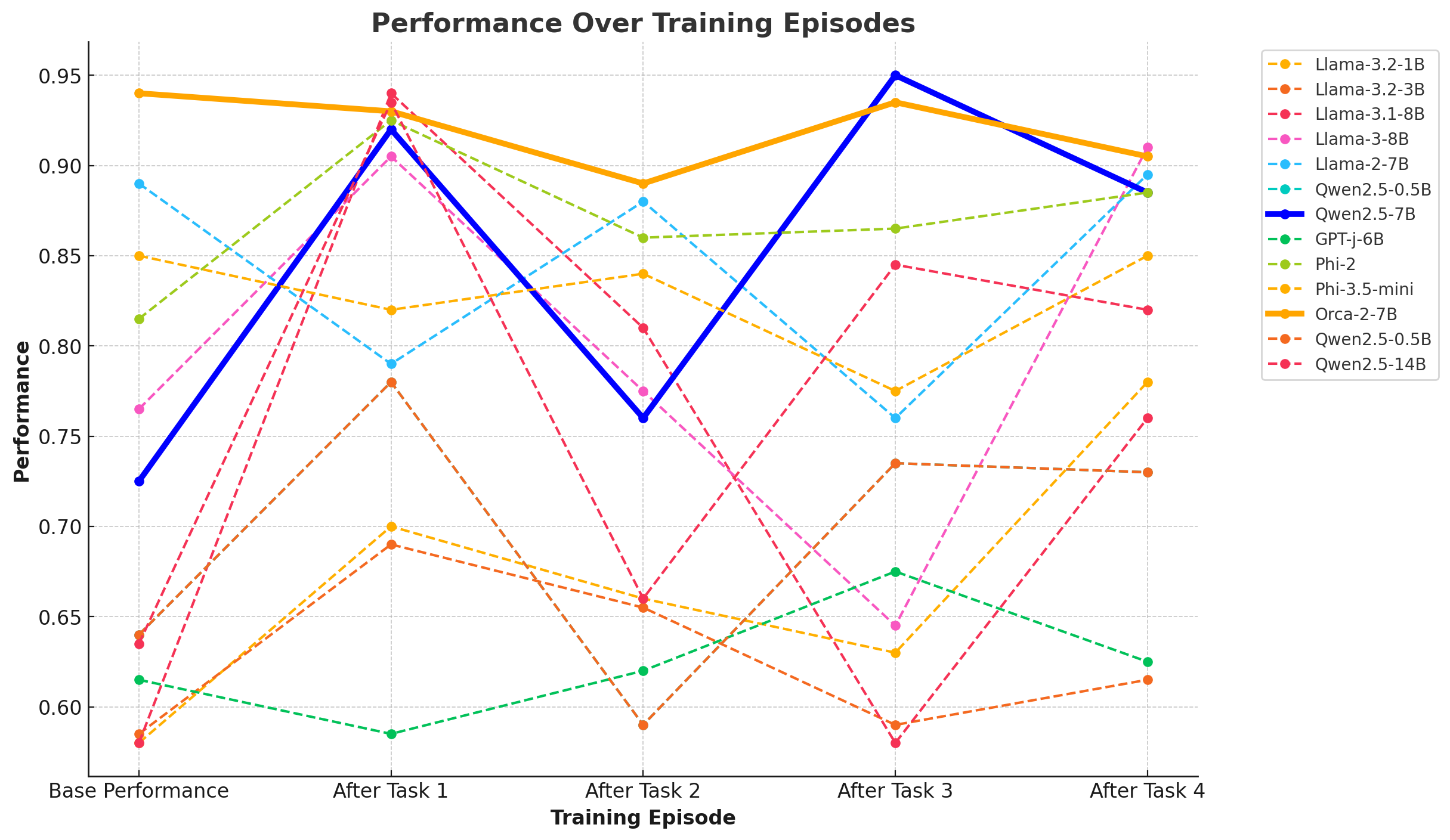}
  \caption{Performance of various models across continual fine-tuning episodes for the task SST2. The solid blue line highlights the model with the highest overall performance, while the solid orange line represents the model with the least amount of forgetting (the smallest drop in performance between tasks). Dashed lines indicate the performance of other models. This diagram illustrates both the learning capacity and retention ability of each model over successive tasks.}
  \label{fig:result}
\end{figure}

Results from continual fine-tuning on the task \textbf{SST2} (Figure~\ref{fig:result}) show \textbf{Qwen2.5-7B} \cite{qwen25} leading in accuracy and learning, while \textbf{Orca-2-7B} \cite{orca} exhibited the least forgetting across tasks.

\begin{figure}[t]
  \includegraphics[width=\columnwidth]{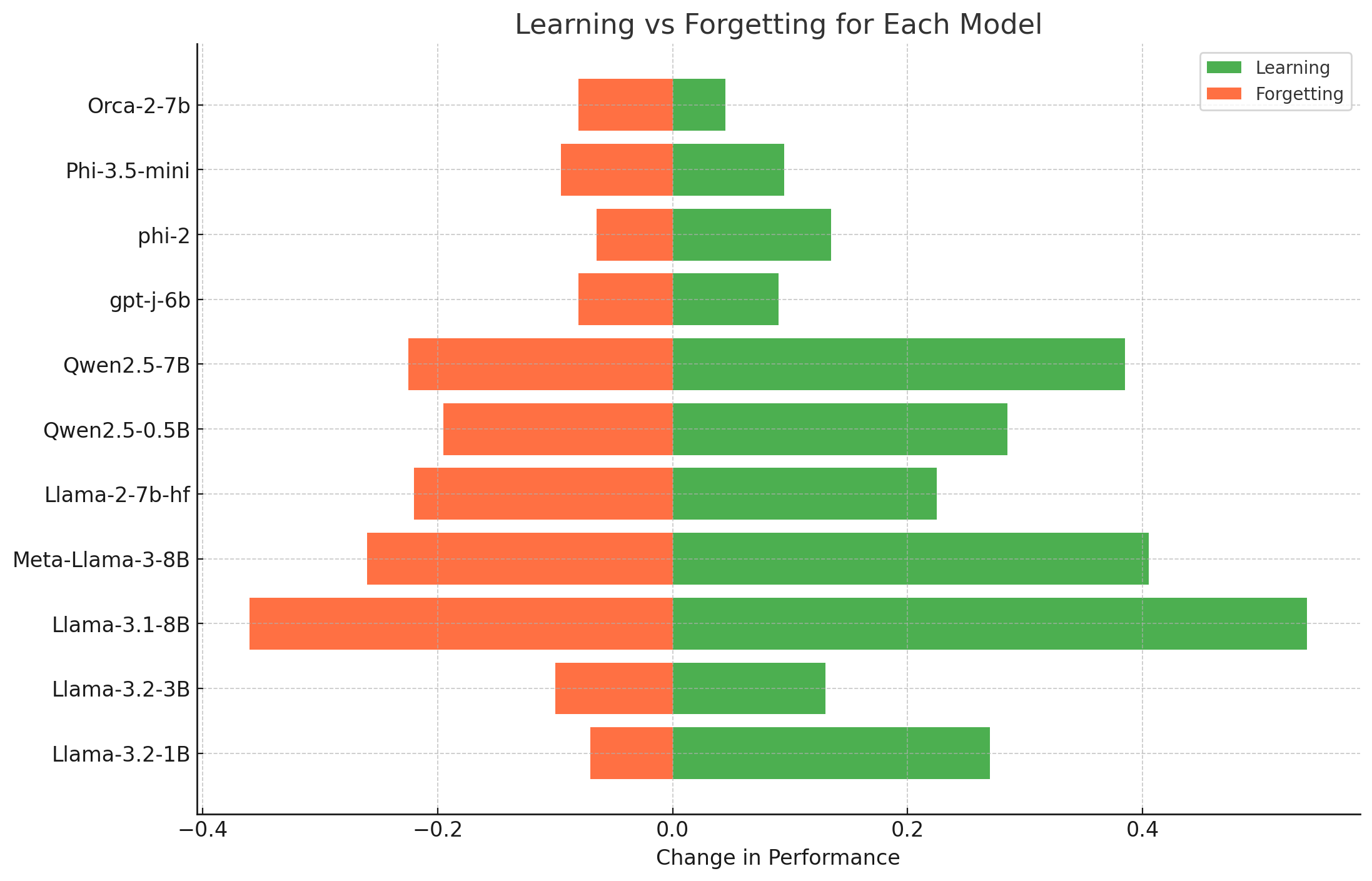}
  \caption{Bar graph displaying model performance for the task SST2 on catastrophic forgetting (reverse) and learning rates, with higher models showing more significant trade-offs. Phi-3.5-mini stands out with minimal forgetting and moderate learning.}
  \label{fig:result2}
\end{figure}

Figure~\ref{fig:result2} shows a clear pattern where larger models, like Qwen2.5-7B and Llama-3.1-8B, exhibit higher learning rates, often at the cost of increased forgetting. In contrast, smaller models, such as Phi-3.5-mini and Phi-2, manage to balance low forgetting with moderate learning gains. This suggests a trade-off between capacity and stability across models.

\section{Conclusion}

This study explored catastrophic forgetting in large language models during sequential fine-tuning on multiple NLU tasks. We found that smaller models like \textbf{Phi-3.5-mini} effectively minimize forgetting while maintaining learning capabilities. Prompt engineering and fine-tuning strategies significantly impact model performance in continual learning settings. Models such as \textbf{Orca-2-7b} and \textbf{Qwen2.5-7B} showed strong learning abilities but varied in forgetting. Careful model selection and tuning can enhance handling multiple tasks without sacrificing accuracy, which is crucial for developing autonomous LLM-based agents. Future work should explore more advanced continual learning techniques to mitigate catastrophic forgetting.

\section{Limitations}

This study focused on models under 10 billion parameters and specific NLU tasks from the GLUE benchmark, so results may not generalize to larger models or other tasks. We only used sequential fine-tuning without exploring other continual learning strategies like memory replay or regularization methods. Relying on prompt engineering may introduce biases affecting performance comparisons. We also didn't consider computational constraints of continual fine-tuning, which could impact practical deployment. Finally, using accuracy as the sole metric might not capture all aspects of model performance.

\bibliography{custom}

\begin{thebibliography}{18}
\providecommand{\natexlab}[1]{#1}

\bibitem[{Bubeck et~al.(2023)Bubeck, Chandrasekaran, Eldan, Gehrke, Horvitz, Kamar, Lee, Lee, Li, Lundberg, and et~al.}]{bubeck2023sparks}
S\'ebastien Bubeck, Varun Chandrasekaran, Ronen Eldan, Johannes Gehrke, Eric Horvitz, Ece Kamar, Peter Lee, Yin~Tat Lee, Yuanzhi Li, Scott~  Lundberg, and et~al. 2023.
\newblock \href {https://arxiv.org/abs/arXiv:2303.12712} {Sparks of artificial general intelligence: Early experiments with gpt-4}.
\newblock ArXiv preprint.

\bibitem[{Chen et~al.(2024)Chen, Zhang, Langrené, and Zhu}]{promptengineering}
Banghao Chen, Zhaofeng Zhang, Nicolas Langrené, and Shengxin Zhu. 2024.
\newblock \href {https://arxiv.org/abs/2310.14735} {Unleashing the potential of prompt engineering in large language models: a comprehensive review}.
\newblock \emph{Preprint}, arXiv:2310.14735.

\bibitem[{Fawi(2024)}]{continual}
Muhammad Fawi. 2024.
\newblock \href {https://doi.org/10.5281/ZENODO.12730055} {Curlora: Stable llm continual fine-tuning and catastrophic forgetting mitigation}.

\bibitem[{Huang et~al.(2024)Huang, Cui, Wang, Yang, Liao, Song, Yao, and Su}]{catastrophicforgettinglarge}
Jianheng Huang, Leyang Cui, Ante Wang, Chengyi Yang, Xinting Liao, Linfeng Song, Junfeng Yao, and Jinsong Su. 2024.
\newblock \href {https://arxiv.org/abs/2403.01244} {Mitigating catastrophic forgetting in large language models with self-synthesized rehearsal}.
\newblock \emph{Preprint}, arXiv:2403.01244.

\bibitem[{Hui et~al.(2024)Hui, Yang, Cui, Yang, Liu, Zhang, Liu, Zhang, Yu, Dang, Yang, Men, Huang, Ren, Ren, Zhou, and Lin}]{qwen25}
Binyuan Hui, Jian Yang, Zeyu Cui, Jiaxi Yang, Dayiheng Liu, Lei Zhang, Tianyu Liu, Jiajun Zhang, Bowen Yu, Kai Dang, An~Yang, Rui Men, Fei Huang, Xingzhang Ren, Xuancheng Ren, Jingren Zhou, and Junyang Lin. 2024.
\newblock \href {https://arxiv.org/abs/2409.12186} {Qwen2.5-coder technical report}.
\newblock \emph{Preprint}, arXiv:2409.12186.

\bibitem[{Kapoor et~al.(2024)Kapoor, Stroebl, Siegel, Nadgir, and Narayanan}]{agents}
Sayash Kapoor, Benedikt Stroebl, Zachary~S. Siegel, Nitya Nadgir, and Arvind Narayanan. 2024.
\newblock \href {https://arxiv.org/abs/2407.01502} {Ai agents that matter}.
\newblock \emph{Preprint}, arXiv:2407.01502.

\bibitem[{Kemker et~al.(2017)Kemker, McClure, Abitino, Hayes, and Kanan}]{kemker2017measuringcatastrophicforgettingneural}
Ronald Kemker, Marc McClure, Angelina Abitino, Tyler Hayes, and Christopher Kanan. 2017.
\newblock \href {https://arxiv.org/abs/1708.02072} {Measuring catastrophic forgetting in neural networks}.
\newblock \emph{Preprint}, arXiv:1708.02072.

\bibitem[{Kirkpatrick et~al.(2017)Kirkpatrick, Pascanu, Rabinowitz, Veness, Desjardins, Rusu, Milan, Quan, Ramalho, Grabska-Barwi{\'n}ska et~al.}]{kirkpatrick2017overcoming}
James Kirkpatrick, Razvan Pascanu, Neil Rabinowitz, Joel Veness, Guillaume Desjardins, Andrei~A Rusu, Kieran Milan, John Quan, Tiago Ramalho, Agnieszka Grabska-Barwi{\'n}ska, et~al. 2017.
\newblock \href {https://doi.org/10.1073/pnas.1611835114} {Overcoming catastrophic forgetting in neural networks}.
\newblock \emph{Proceedings of the National Academy of Sciences}, 114(13):3521--3526.

\bibitem[{Llama~Team(2024)}]{llama3.2}
AI~Llama~Team. 2024.
\newblock \href {https://arxiv.org/abs/2407.21783} {The llama 3 herd of models}.
\newblock \emph{Preprint}, arXiv:2407.21783.

\bibitem[{Lopez-Paz and Ranzato(2017)}]{lopez2017gradient}
David Lopez-Paz and Marc’Aurelio Ranzato. 2017.
\newblock Gradient episodic memory for continual learning.
\newblock In \emph{Advances in Neural Information Processing Systems}, volume~30, pages 6467--6476.

\bibitem[{Luo et~al.(2024)Luo, Yang, Meng, Li, Zhou, and Zhang}]{empiricalstudycatastrophicforgetting}
Yun Luo, Zhen Yang, Fandong Meng, Yafu Li, Jie Zhou, and Yue Zhang. 2024.
\newblock \href {https://arxiv.org/abs/2308.08747} {An empirical study of catastrophic forgetting in large language models during continual fine-tuning}.
\newblock \emph{Preprint}, arXiv:2308.08747.

\bibitem[{McCloskey and Cohen(1989)}]{mccloskey1989catastrophic}
Michael McCloskey and Neal~J. Cohen. 1989.
\newblock \href {https://doi.org/10.1016/S0079-7421(08)60536-8} {Catastrophic interference in connectionist networks: The sequential learning problem}.
\newblock \emph{Psychology of Learning and Motivation}, 24:109--165.

\bibitem[{Microsoft(2023)}]{orca}
Microsoft. 2023.
\newblock \href {https://arxiv.org/abs/2311.11045} {Orca 2: Teaching small language models how to reason}.
\newblock \emph{Preprint}, arXiv:2311.11045.

\bibitem[{Microsoft(2024)}]{phi}
Microsoft. 2024.
\newblock \href {https://arxiv.org/abs/2404.14219} {Phi-3 technical report: A highly capable language model locally on your phone}.
\newblock \emph{Preprint}, arXiv:2404.14219.

\bibitem[{Scialom et~al.(2022)Scialom, Chakrabarty, and Muresan}]{tuhin}
Thomas Scialom, Tuhin Chakrabarty, and Smaranda Muresan. 2022.
\newblock Fine-tuned language models are continual learners.
\newblock \emph{arXiv preprint arXiv:2210.05653}.

\bibitem[{Vaswani et~al.(2017)Vaswani, Shazeer, Parmar, Uszkoreit, Jones, Gomez, Kaiser, and Polosukhin}]{attention}
Ashish Vaswani, Noam Shazeer, Niki Parmar, Jakob Uszkoreit, Llion Jones, Aidan~N. Gomez, Lukasz Kaiser, and Illia Polosukhin. 2017.
\newblock \href {https://arxiv.org/abs/1706.03762} {Attention is all you need}.
\newblock \emph{CoRR}, abs/1706.03762.

\bibitem[{Wang et~al.(2018)Wang, Singh, Michael, Hill, Levy, and Bowman}]{glue}
Alex Wang, Amanpreet Singh, Julian Michael, Felix Hill, Omer Levy, and Samuel~R. Bowman. 2018.
\newblock \href {https://arxiv.org/abs/1804.07461} {{GLUE:} {A} multi-task benchmark and analysis platform for natural language understanding}.
\newblock \emph{CoRR}, abs/1804.07461.

\bibitem[{Wang(2021)}]{mesh-transformer-jax}
Ben Wang. 2021.
\newblock {Mesh-Transformer-JAX: Model-Parallel Implementation of Transformer Language Model with JAX}.
\newblock \url{https://github.com/kingoflolz/mesh-transformer-jax}.

\end{thebibliography}

\appendix

\section{Example Appendix}
\label{sec:appendix}

This is an appendix.

\end{document}